\documentclass{article}
\usepackage[nonatbib]{neurips_2020}
\usepackage[multiple]{footmisc}
\nolinenumbers

\usepackage[utf8]{inputenc} 
\usepackage[T1]{fontenc}    
\usepackage{hyperref}       
\usepackage{url}            
\usepackage{booktabs}       
\usepackage{amsfonts}       
\usepackage{nicefrac}       
\usepackage{microtype}      
\usepackage{graphicx}
\usepackage{amsmath} 
\usepackage{multirow}
\usepackage{graphicx}
\newcommand{\tabref}[1]{Table~\ref{#1}}
\newcommand{\figureref}[1]{Figure~\ref{#1}}

\title{Prototype-oriented Unsupervised Change Detection\\for Disaster Management}

\author{Youngtack Oh, \hspace{3mm} Minseok Seo, \hspace{3mm}Doyi Kim, \hspace{3mm}Junghoon Seo \\
SI Analytics\\
70, Yuseong-daero 1689beon-gil, Yuseong-gu, Daejeon, Republic of Korea \\
\texttt{\{ytoh96,minseok.seo,doyikim,jhseo\}@si-analytics.ai} \\
}
\begin{document}

\maketitle

\begin{abstract}
Climate change has led to an increased frequency of natural disasters such as floods and cyclones. This emphasizes the importance of effective disaster monitoring. In response, the remote sensing community has explored change detection methods. These methods are primarily categorized into supervised techniques, which yield precise results but come with high labeling costs, and unsupervised techniques, which eliminate the need for labeling but involve intricate hyperparameter tuning. To address these challenges, we propose a novel unsupervised change detection method named \textit{Prototype-oriented Unsupervised Change Detection for Disaster Management} (PUCD). PUCD captures changes by comparing features from pre-event, post-event, and prototype-oriented change synthesis images via a foundational model, and refines results using the \textit{Segment Anything Model} (SAM). Although PUCD is an unsupervised change detection, it does not require complex hyperparameter tuning. We evaluate PUCD framework on the LEVIR-Extension dataset and the disaster dataset and it achieves state-of-the-art performance compared to other methods on the LEVIR-Extension dataset.
\end{abstract}

\section{Introduction}
Climate change is leading to more extreme weather events~\cite{ipcc2023}. With the acceleration of global warming, damage from climate-related disasters such as floods and tropical cyclones is escalating~\cite{bhatia2019recent, lin2020tropical}. As a result, there is heightened emphasis on the critical role of disaster management in both prevention and response~\cite{sun2020applications}. Employing remote sensing data for disaster management proves efficient as it enables the monitoring of large areas~\cite{fan2021disaster, yu2018big}. However, even though remote sensing is efficient,  the large area affected by natural disasters makes it difficult for humans to analyze~\cite{li2019unsupervised}. Addressing these challenges, unsupervised change detection (UCD) methods have been introduced, utilizing change vector analysis (CVA) of pre- and post-disaster images~\cite{liu2019review, pcakmeans, irmad, dcva, sfa, 5196726}. While UCD allows for rapid responses, it necessitates the adjustment of numerous hyperparameters. This is because UCD uses pixel statistical differences such as brightness, color, and clarity of the image, so it is difficult to generalize differences across various disaster types and regions. On the other hand, supervised change detection \cite{9355573,bandara2022transformer,zhang2023asymmetric, seo2023self} can be used without the need to adjust hyperparameters. However, it is difficult to build large-scale datasets because disaster situations are rare~\cite{Hamaguchi_2019_CVPR}. Specifying and labeling targets of interest also requires a significant cost~\cite{yao2016semantic}. In addition, these models cannot respond to disasters that are not trained.
\begin{figure*}[t!]
    \centering
    \includegraphics[width=1.0\columnwidth]{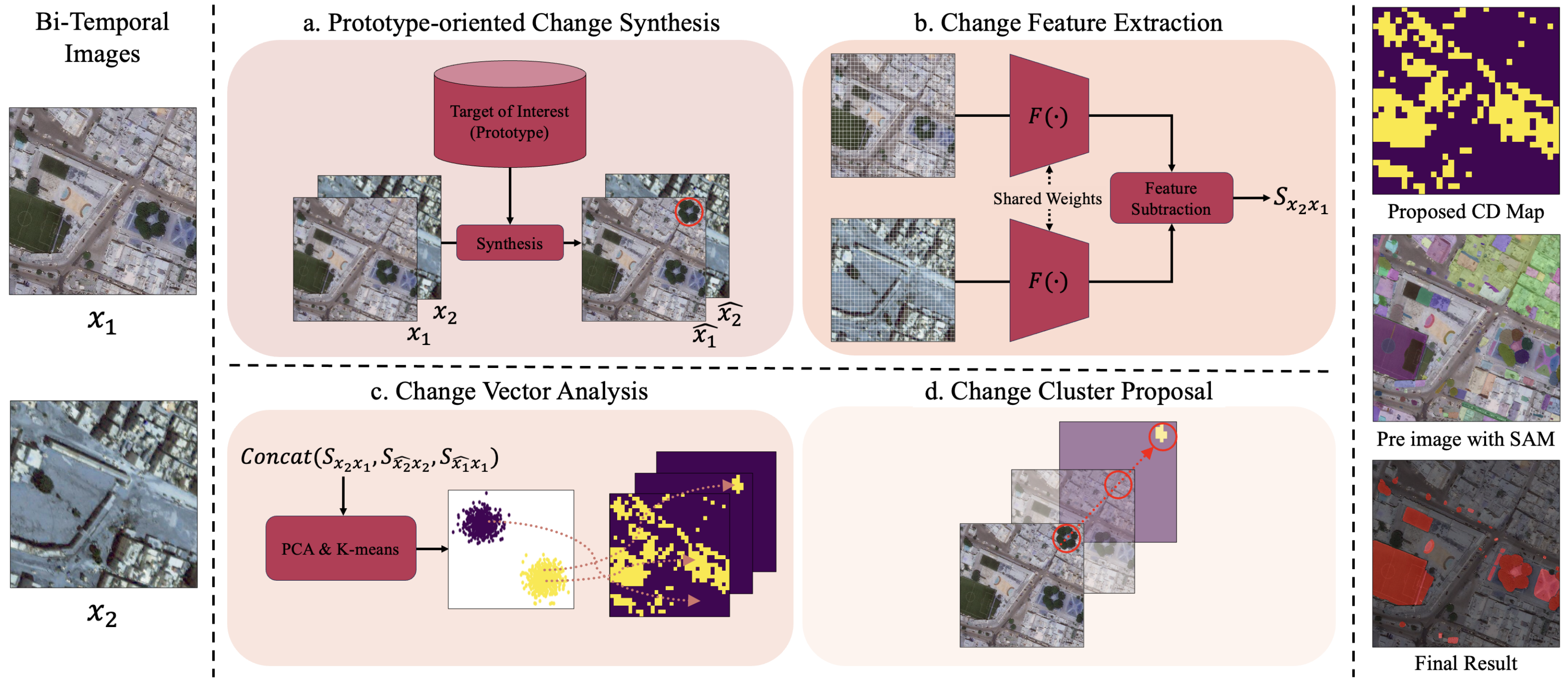}
    \caption{Overview of our PUCD framework. The framework receives a bi-temporal pair of images: the pre-disaster image \( x_{1} \) and the post-disaster image \( x_{2} \). \textbf{(a)} A prototype is then synthesized as \( \hat{x}_{1}\) and \( \hat{x}_{2}\) . \textbf{(b)} Subsequent steps involve feature extraction and subtraction them output \(S_{x_{2}x_{1}} \), \(S_{\hat{x}_{1}x_{1}} \) and \(S_{\hat{x}_{2}x_{2}} \). \textbf{(c)} change vector creation through CVA. \textbf{(d)} A change map emerges from change clustering. The final output is a fusion of this map and the results from the SAM.}
    \label{fig:fig1_final_real_label}
\end{figure*}

In this paper, we introduce a \textit{Prototype-oriented Unsupervised Change Detection for Disaster Management} (PUCD), illustrated as \figureref{fig:fig1_final_real_label}. This approach designates a target-of-interest as a prototype and detects changes relative to it. PUCD uses the DINOv2 foundational model to extract features representing the context and structure of pre-, post-event imageries and prototypes. Following this, CVA is applied to the features of the pre-, post-event images and the prototype to identify changed regions. The preliminary coarse output is then enhanced using the \textit{segment anything} technique \cite{kirillov2023segment}.
To evaluate the PUCD framework, we conducted qualitative assessments on events like the Libyan flood, the United States hurricane, and the Japan tsunami.
Subsequent quantitative evaluations were executed using the LEVIR-Extension dataset.
Our experimental results showed that PUCD is qualitatively and quantitatively efficient.
\section{Method}
In this section, we provide a detailed explanation of the PUCD framework.
Initially, we summarize the components constituting PUCD, specifically the DINOv2 and the \textit{segment anything} model.
Subsequently, we delve into the contents corresponding to ~\figureref{fig:fig1_final_real_label}-(a), (b), (c), and (d) in sequence.
\subsection{Preliminaries}
\paragraph{DINOv2} DINOv2~\cite{oquab2023dinov2} is a foundational model rooted in representation learning, widely used in various computer vision tasks.
It utilizes the Vision Transformer (ViT) architecture~\cite{dosovitskiy2021an,he2020momentum} and is trained on a comprehensive compilation of diverse datasets~\cite{oquab2023dinov2}.
DINOv2 is trained using contrastive learning, assessing the similarity between objects in two images based on their context and shape \cite{caron2021emerging,oquab2023dinov2}.
Consequently, DINOv2 is not only capable of capturing changes in context and shape but also exhibits robustness to stylistic change.
This capability underpins our rationale for employing DINOv2, particularly for UCD tasks that necessitate detecting changes in context and shape.
\paragraph{Segment Anything} The Segment Anything Model (SAM)~\cite{kirillov2023segment}, built upon the MaskFormer architecture \cite{cheng2021per}, is a universal segmentation solution.
It is recognized as a foundational model in the segmentation field, having been trained on a massive instance segmentation dataset comprised of 1 billion masks and 11 million images.
SAM consistently produces fine-grained segmentation results across diverse applications, ranging from satellite imagery and medical imaging to autonomous driving~\cite{mediSAM1, MediSAM2, TrackSAM, RSSAM}.
Given its capabilities, SAM holds significant promise for enhancing the precision of change detection in the remote sensing domain.
\subsection{PUCD framework}
\paragraph{Prototype-oriented Change Synthesis}
Satellite imagery exhibits significant variation in resolution, and even within the same resolution category, the visual representation of target objects can substantially differ~\cite{theng2006automatic}. 
Given that disasters can occur globally, satellite images with varying resolutions play a crucial role in disaster management.
However, the context and structural attributes of objects differ depending on the region.
Addressing this complexity necessitates a dynamic change threshold, which considers both satellite resolution and specific object attributes.
To meet this need, we introduce a \textit{prototype}-oriented change detection.

A \textit{prototype} (\( p \)) refers to a predefined single-sample image representing a target of interest, such as a house or sports field.
By storing these images in advance, the system can define changes with respect to the prototype, offering a structured method to handle different imaging contexts. 
The PUCD framework, while allowing for the specification of a prototype, remains adaptable enough to operate without manually defined one.
Notably, if \( p \) is not provided from user's choices, an object from either \( x_{1} \) or \( x_{2} \) is randomly selected as \( p \) from results of SAM, streamlining change detection across various regions or targets and obviating exhaustive hyperparameter tuning. 
Note that the prototype aids in the assignment of change clusters, discussed in the subsequent section.
\paragraph{Change Feature Extraction} %
\label{sec:algo1}
Given the bi-temporal images \( (x_{1}, x_{2}) \) and the images  generated through prototype-oriented change synthesis \( (\hat{x}_{1}, \hat{x}_{2}) \), PUCD extracts the corresponding features.
Specifically, PUCD divides these images into multiple patches, each of size \( 14 \times 14 \).
Each patch is then processed by DINOv2 to extract its features.
Essentially, PUCD derives features of context and structure from a confined \( 14 \times 14 \) area for image comparison.
This approach is especially advantageous in situations where the same area of interest is captured in bi-temporal images from satellites, but spatial discrepancies, referred to as co-registration errors, exist between these images \cite{9740657, tuia2016domain}.
\begin{figure*}[t!]
    \centering
    \includegraphics[width=\columnwidth]{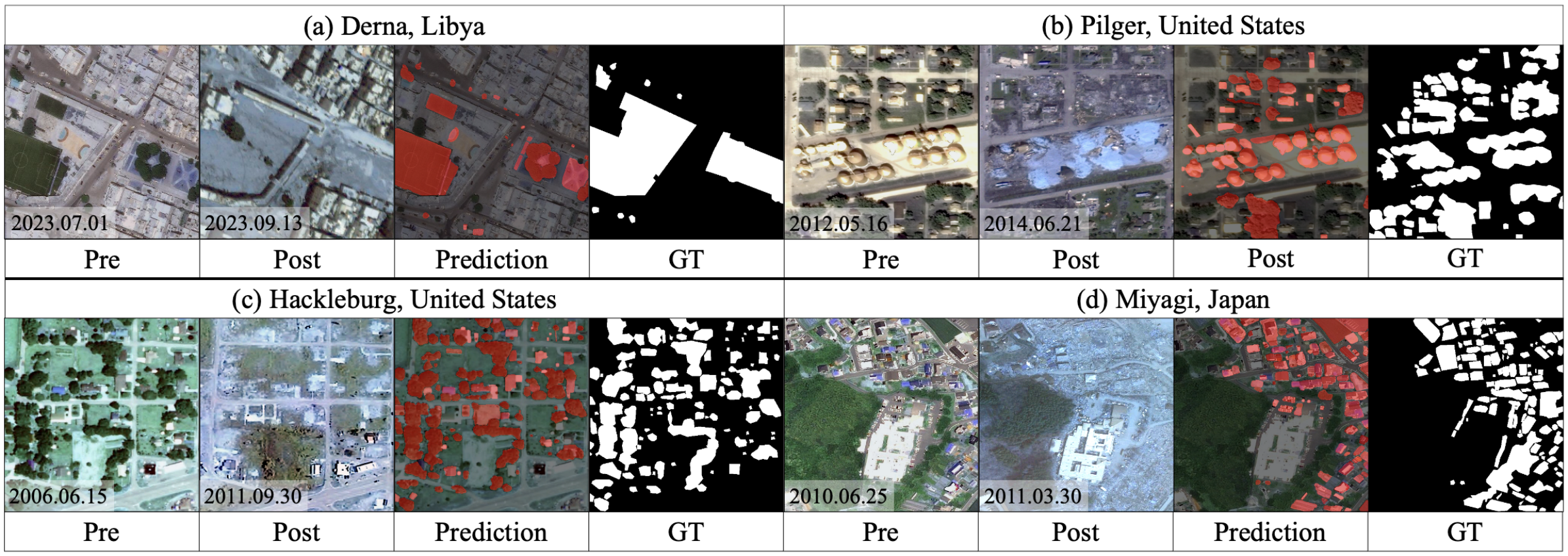}
    \caption{Results of qualitative analysis of PUCD. Note that all qualitative experiments were performed by selecting a random object from images \( x_{1} \) and \( x_{2} \) using the \textit{SAM} model, rather than manually selecting a specific prototype \( p \).}
    \label{fig:fig2_real5_final}
\end{figure*}
\paragraph{Change Vector Analysis \& Change Cluster proposal}
The CVA of PUCD identifies change features using the \( \hat{x}_{1}\) and \( \hat{x}_{2} \), in conjunction with the \( x_{1} \) and \( x_{2} \).
Specifically, \( S_{x_{2}x_{1}} \), \( S_{\hat{x}_{1}x_{1}} \), and \( S_{\hat{x}_{2}x_{2}} \) represent the feature differences between \( x_{2} \) and \( x_{1} \), the \(\hat{x}_{1}\) and \( x_{1} \), and the  \(\hat{x}_{2}\) and \( x_{2} \), respectively.
After concatenating the features $S_{x_{2}x_{1}}$, $S_{\hat{x}_{1}x_{1}}$, and $S_{\hat{x}_{2}x_{2}}$, they undergo dimensionality reduction using Principal Component Analysis (PCA)~\cite{jollife2016principal}. 
Beyond its primary role in reducing dimensions, it accentuates pertinent features while diminishing data noise, ensuring the feature space concentrates on the most crucial variances.
Once significant features have been extracted via PCA, K-means clustering categorizes these features, distinguishing changes from unchanges.
Within this scheme, the cluster that contains the prototype is identified as the change cluster.
\paragraph{Refinement with SAM Model}
A single pixel within the feature map from DINOv2 corresponds to a \( 14 \times 14 \) area in the original image.
Consequently, detected changes often manifest over an excessively larger region than a size of \( 14 \times 14 \).
To address this issue, we utilize the SAM model. Upon obtaining a coarse change map \( S_{\text{map}} \), PUCD refines the \( x_{1} \) image. Specifically, only those objects with an overlap exceeding 70\% with the semantic SAM segmentation map are retained.

\section{Experiment Results}
In our experiments, we employed the DINOv2 ViT-L/14 distilled model~\cite{oquab2023dinov2} for DINOv2 and utilized the SAM-H~\cite{kirillov2023segment} model for SAM. Within the PUCD framework, the number of PCA components was set to 1, and for k-means clustering, the value of \(k\) was set to 2(change and unchange).

To assess the efficacy of the PUCD framework, we conducted a quantitative evaluation using the \textit{LEVIR-Extension} dataset. Additionally, qualitative evaluations were carried out on events such as the Libyan flood, the hurricane in the United States, and the tsunami in Japan.
The LEVIR-CD dataset~\cite{rs12101662} comprises 637 bi-temporal pairs of high spatial resolution (HSR) optical images with labels indicating building changes.
Each image has a spatial size of 1,024$\times$1,024 pixels and a ground sample distance (GSD) of 0.5 $m$.
To capture all major changes in the LEVIR-CD dataset, we added labels for roads, parking lots, lakes, and other features. This augmented dataset, named \textit{LEVIR-Extension}, will be available online shortly. The pre/post-disaster dataset for qualitative results was sourced from MAXAR's WorldView-2 and GeoEye-1 imagery, with GSDs of $0.46m$ and $0.25m$, respectively.
\begin{table*}[t!]
\centering
\caption{Performance comparison of unsupervised change detection methods (CVA, IRMAD, PCAKmeans, SFA, DCVA) that require co-registered bi-temporal images on the LEVIR-Extension dataset. Please note that the best scores are indicated in \textbf{bold}, and the second-best scores are \underline{underlined}.}
\label{tab:tab1}
\resizebox{1.0\columnwidth}{!}{%
\begin{tabular}{cc|ccccc}
\hline \hline
\multirow{2}{*}{\textbf{Method}} & \multirow{2}{*}{\textbf{Param(M) 
}} & \multicolumn{5}{c}{\textbf{LEVIR-Extension}}  \\ \cline{3-7} 
&             & Pre. (0/1)      & Rec. (0/1)     & F1 (0/1)    & IoU (0/1)    & ACC       \\ \hline
CVA~\cite{cva}&  -&45.4/\textbf{60.9}&92.7/9.3&60.9/16.2&43.8/8.8&46.7 \\
IRMAD~\cite{irmad}&  -&73.6/46.3&93.7/13.9&82.5/21.4&70.1/12.0&71.3\\
PCAKmeans~\cite{pcakmeans}&  -&63.7/43.6&92.5/9.9&75.4/16.2&60.6/8.8&62.0\\
SFA~\cite{sfa}& -&76.9/\underline{48.0}&\underline{94.1}/16.1&84.7/24.1&73.4/13.7&74.5\\
DCVA~\cite{dcva}& 16.15 &94.1/14.3&92.3/18.2&\underline{93.2}/16.0&\underline{87.2}/8.7&\underline{87.41}\\ \hline
SAM-CD &  632 & 92.7/10.9 & 66.9/43.7 & 77.7/17.4 & 63.6/9.5& 65.0 \\
PUCD w/o  SAM & 932 & \textbf{97.4}/23.1 & 75.8/\textbf{78.6} & 85.2/\underline{35.6} & 74.3/\underline{21.7} & 76.0 \\
PUCD w/ SAM & 932 & \underline{95.0}/47.7 & \textbf{95.3}/\underline{46.2} & \textbf{95.1}/\textbf{46.8} & \textbf{90.8}/\textbf{30.7} & \textbf{91.1} \\ \hline 
\hline
\end{tabular}%
}
\end{table*}
%
\paragraph{Qualitative Results}
%
%
~\figureref{fig:fig2_real5_final} showcases the change detection results for various disasters, including floods, hurricanes, and typhoons.
The figure clearly demonstrates PUCD's ability to detect signs of damage, such as destroyed buildings, flooded areas, and damaged trees.
Remarkably, these results were achieved without any training procedure on the target dataset, highlighting its capability as an UCD model and its adaptability across different regions and disaster types.
%
\paragraph{Quantitative  Results}
~\tabref{tab:tab1} presents the performance comparison of PUCD with other UCD methodologies using the LEVIR-Extension dataset.
The table results reveal that PUCD consistently outperforms existing methods, achieving state-of-the-art results with a significant margin.
These findings underscore the PUCD framework's potential in the domain of UCD.
\section{Conclusion}
In this paper, we introduced the Prototype-oriented Unsupervised Change Detection (PUCD), which delivers robust performance regardless of satellite image resolution or the specific subject of interest.
By leveraging features from both pre-event and post-event imagery and incorporating prototypes, followed by refining results through the \textit{Segment Anything} technique, PUCD achieves superior accuracy over the other unsupervised change detectors.
Qualitative assessments across various disaster types further attest to its robust performance across different regions.
We foresee PUCD becoming a cornerstone tool in future disaster management efforts.
\clearpage
{\small
\bibliographystyle{plain}
\bibliography{egbib}
}
\end{document}